\documentclass{article}
\usepackage{iclr2017_conference,times}
\usepackage{url}
\usepackage{latexsym}
\usepackage{epigraph}
\usepackage[natbib,maxcitenames=2,maxbibnames=99,style=authoryear-comp,dashed=false,sorting=nyt,mincitenames=1,backend=bibtex,safeinputenc,%
    texencoding=utf8,bibencoding=utf8,uniquename=false,uniquelist=false,%
    useprefix=true,%
    doi=false,isbn=false,url=false]{biblatex}
\usepackage{booktabs}
\usepackage{enumitem}
\usepackage{color}
\usepackage{graphicx}
\usepackage{caption}
\usepackage{subcaption}
\usepackage{amsthm}
\usepackage{amsfonts}
\usepackage{siunitx}
\usepackage{tabu}
\usepackage[hidelinks]{hyperref}
\usepackage{pdfcomment}
\usepackage[capitalise]{cleveref}
\theoremstyle{definition}
\newtheorem{definition}{Definition}[]
\definecolor{maluubablue}{HTML}{104586}

\newenvironment{example}{\quotation}{\endquotation}
\newcommand{\ndialogues}{1369}
\newcommand\wizardcell[1]{\multicolumn{1}{l}{\color{maluubablue}#1}}

\title{Frames: A Corpus for Adding Memory to Goal-Oriented Dialogue Systems}

\author{Layla El Asri$^*$ \and \bf Hannes Schulz$^*$ \and \bf Shikhar Sharma$^*$ \and \bf Jeremie Zumer$^*$
    \And Justin Harris \and \bf Emery Fine \and \bf Rahul Mehrotra \and \bf Kaheer Suleman
    \vspace*{3mm}
    \AND Maluuba Research \AND Montreal, Canada \AND \texttt{first.last@microsoft.com}
    \AND \bf $^*$these authors contributed equally}

\begin{document}
\maketitle
\begin{abstract}

This paper presents the \emph{Frames} dataset\footnote{Frames is available at \url{http://datasets.maluuba.com/Frames}.}, a corpus of \ndialogues{} human-human dialogues with an average of 15 turns per dialogue. We developed this dataset to study the role of memory in goal-oriented dialogue systems. Based on Frames, we introduce a task called \emph{frame tracking}, which extends state tracking to a setting where several states are tracked simultaneously. We propose a baseline model for this task. We show that Frames can also be used to study memory in dialogue management and information presentation through natural language generation.

\end{abstract}

\section{Introduction}
Goal-oriented, information-retrieving dialogue systems have traditionally been designed to help users find items in a database given a certain set of constraints \citep{Singh:02,Raux:03,ElAsri:14,classicFinal}. For instance, the \textit{LET'S GO} dialogue system finds a bus schedule given a bus number and a location \citep{Raux:03}.

These systems model dialogue as a sequential process: the system asks for constraints until it can query the database and return a few results to the user. Then, the user can ask for more information about a given result or ask for other possibilities. If the user wants to know about database items corresponding to a different set of constraints (\textit{e.g.}, another bus line), then these constraints simply overwrite the previous ones. As a consequence, users can neither compare results corresponding to different constraints, nor go back-and-forth between results. 

We can assume users in the bus domain know exactly what they want. In contrast, user studies in e-commerce have shown that several information-seeking behaviours are encountered: users may come with a very well defined item in mind, but they may also visit an e-commerce website with the intent to compare items and explore different possibilities \citep{Moe:01}. Supporting this kind of decision-making process in conversational systems implies adding \emph{memory}. Memory is necessary to track different items or preferences set by the user during the dialogue. For instance, consider product comparisons. If a user wants to compare different items using a dialogue system, then this system should be able to separately recall properties pertaining to each item.

This paper presents the \emph{Frames} dataset, which comprises dialogues that require this type of memory. Frames is a corpus of \ndialogues{} human-human dialogues collected in a Wizard-of-Oz (WOz) setting -- \textit{i.e.},~users were connected to humans, whom we refer to as \textit{the wizards}, who were assuming the role of the dialogue system. The wizards had access to a database of vacation packages containing round-trip flights and a hotel. The users were asked to find packages based on a few constraints such as a destination and a budget. 

In order to test the memory capabilities of conversational agents, we formalize a new task called \emph{frame tracking}. In frame tracking, the agent must simultaneously track multiple semantic frames throughout the dialogue. For example, two frames would be constructed and recalled while comparing two products -- each containing the properties of a specific item. Frame tracking is a generalization of the state tracking task \citep{Henderson:15}. In state tracking, all the information summarizing a dialogue history is compressed into one semantic frame. In contrast, several frames are kept in memory during frame tracking, with each frame corresponding to a particular context, \textit{e.g.},~one or more vacation packages.

Another important property of human dialogue that we want to study with the Frames dataset is how to provide the user with information on the database. When a set of user constraints leads to no results, users would benefit from knowing that relaxing a given constraint (\textit{e.g.}, increasing the budget by a reasonable amount) would lead to results instead of navigating the database blindly. This recommendation behaviour is in accordance with Grice's cooperative principle \citep{Grice:89}: ``Make your contribution as informative as is required (for the current purposes of the exchange)''. We study this by including suggestions when the database returns no results for a given user query.

This paper describes the Frames dataset in detail, formally defines the frame tracking task, and provides a baseline model for frame tracking. The next section discusses motivation for the Frames dataset. \cref{sec:data_collection} explains the data collection process and \Cref{corpus_characteristics} describes the dataset in detail. We describe the annotation scheme in \cref{annotation}. In \cref{sec:challenges}, we identify the main research topics of the corpus and formalize the frame tracking task. The dialogue data format is described in \cref{dataset_format}. \Cref{baseline} proposes a baseline model for frame tracking. Finally, we conclude in \cref{conclusion} and suggest directions for future work.

\section{Motivation}
Much work has focused on spoken dialogue \citep{Walker:98,Lemon:07,Williams:16}, since spoken dialogue systems are useful in many settings, including in hands-free environments such as cars \citep{Lemon:06}. A generation of voice assistants -- such as SIRI, Cortana, and Google Voice -- have popularized spoken dialogue systems. More recently, users have become familiar with \textit{chatbots}. Many platforms for deploying chatbots are now available, such as Facebook Messenger, Slack, or Kik. Text offers advantages over voice such as privacy and the ability to avoid bad speech recognition in noisy environments. Chatbots provide a welcome alternative to downloading and installing applications, and make a lot of sense for everyday services such as ordering a cab or knowing the weather. Chatbots have been proposed for tasks that one would traditionally perform through Graphical User Interfaces (GUIs). For instance, many chatbots for booking a flight are now entering the market.

In most cases, as with current voice-based assistants, the conversation with a chatbot is very limited: asking for the weather and ordering a cab are accomplished with simple, sequential slot-filling. These tasks have in common the fact that in both cases the user knows exactly what she wants, \textit{i.e.},~the destination for the cab or the city for the weather. Booking a flight is a bit different. Flight booking requires specifying many parameters, and these are usually determined during the search process\footnote{One can easily imagine a user changing from economy to business class if the price difference is small.}. Technically, finding a weather forecast is only about reading a database: the task is to form a complete database query and then to verbalise the result to the user. The user might start with very few constraints and then refine her query given the database results. In the case of booking a flight, there is a decision-making process requiring comparison and backtracking.

GUIs are not optimal on many levels when it comes to helping users through this decision-making process. A first point of friction is the limited visual space. Consider the example of e-commerce websites. A user is very likely to compare different options before picking an item to buy. This often results in a large number of open browser tabs among which the user must navigate. In order to avoid this situation, some websites provide a comparator that can be used to display several items on a single page. However, this option is not optimal for hierarchical objects such as vacation packages because these objects have global properties (dates of the trips) but are also composed of different modules (flights and hotel) which have different properties (\textit{e.g.}, seat class and hotel category). Optimally, the user should be able to define the properties for the different modules while being able to compare items corresponding to each set of properties. A text interface could complement a GUI by offering this flexibility while remembering the properties mentioned by the user and displaying comparisons when asked. 

We propose the Frames dataset to support work on text-based conversational agents which help a user make a decision. The decision-making process is tightly coupled with the notion of memory. Indeed, if a user intends to compare different options in the course of defining the options, the system should follow the user's path and remember every option. In this paper, we formalize this aspect of conversation in the \emph{frame tracking} task. This task is the main challenge of the Frames corpus, and \cref{sec:challenges} describes it in detail.

\section{Data Collection}
\label{sec:data_collection}
We collected data over a period of 20 days and with 12 participants. To increase variation in the dialogues, 8 participants were hired for only one week each, and one participant was hired for one day. The three remaining participants participated in the entire data collection. The participants were paired up for each dialogue and interacted through a chat interface.

\subsection{Wizard-Of-Oz Data Collection}
Data collection for goal-oriented dialogue is challenging. To control the data such that specific aspects of the problem can be studied, it is common to collect dialogues using an automated system. This requires, \textit{e.g.}, a natural language understanding module that already performs well on the task, which implies possession of in-domain data or the ability to generate it \citep{Raux:03,Henderson:14}. Another possibility, which permits even greater control, is to generate dialogues using a rule-based system \citep{Bordes:16}. These approaches are useful for studying specific modules and analysing the behaviour of different architectures. However, it is costly to generate new dialogues for each experiment and skills acquired on artificial data are not directly usable in real settings because of natural language understanding noise \citep{Bordes:16}.

The Wizard-of-Oz (WOz) approach offers a powerful alternative \citep{Kelley:84,Rieser:05,Wen:16}. In WOz data collection, one participant (the wizard) plays the role of the dialogue system. The wizard has access to a search interface connected to the database. She receives the user's input in text form and decides what to say next. This does not require preexisting dialogue system components, except potentially automatic speech recognition for transcribing the user's inputs. Dialogues collected in WOz settings can be used for studying and developing every part of a dialogue system, from language understanding to language generation. They are also essential for offline training of end-to-end dialogue systems \citep{Wen:16,Bordes:16} on different domains, which may reduce costs from handcrafting new systems for each domain. 

WOz dialogues also have the considerable advantage of exhibiting realistic behaviours that cannot be supported by current (end-to-end or not) architectures. Since there is no dialogue system that incorporates the type of memory that we want to study with this dataset, we need to work directly on human-human dialogues. Our setting is a bit different from the usual WOz setting because, in our case, the users did not think they were interacting with a dialogue system but instead knew that they were talking to a human-being. We made the choice not to give templated answers to the wizards because, apart from studying memory, we also want to study information presentation and dialogue management. We have chosen to work on text-based dialogues because this allows a more controlled wizard behaviour, obviates handling time-sensitive turn-taking and speech recognition noise, and allows studying more complex dialogue flows.

\subsection{Task Templates and Instructions}
Dialogues were performed on Slack\footnote{\url{www.slack.com}}. We deployed a Slack bot named \textit{wozbot} to pair up participants and record conversations. The participants in the user role indicated when they were available for a new dialogue through this bot. They were then assigned to an available wizard and received a new task. The tasks were built from templates such as the following:
\begin{example}
  ``Find a vacation between \texttt{[START\_DATE]} and \texttt{[END\_DATE]} for \texttt{[NUM\_ADULTS]}
  adults and \texttt{[NUM\_CHILDREN]} kids. You leave from \texttt{[ORIGIN\_CITY]}. You are
  travelling on a budget and you would like to spend at most \$\texttt{[BUDGET]}.''
\end{example}
Each template had a probability of success. The tasks were generated by drawing values (\textit{e.g.}, \texttt{BUDGET}) from the database. The generated tasks were then added to a pool. The values were drawn in order to comply with the template's probability of success. For example, if 20 tasks were generated at probability 0.5, about 10 tasks would be generated with successful database queries and the other 10 would be generated so the database returned no results for the constraints. This mechanism allowed us to emulate cases when a user would not find anything meeting her constraints. If a task was unsuccessful, the user either ended the dialogue or got an alternate task such as:
\begin{example}
  ``If nothing matches your constraints, try increasing your budget by \$200.''
\end{example}
We wrote 38 templates. 14 templates were generic such as the one presented above and the other 24 were written to encourage more role-playing from users. One example is:
\begin{example}
  ``Pokemon are now a worldwide currency. You are the best Pokemon hunter in the
  world. You have caught them all except for one Pokemon worth a fortune: Mewtwo.
  You heard it was spotted somewhere in \texttt{[DESTINATION\_CITY]} and
  \texttt{[DESTINATION\_CITY]}. You want to visit one of these cities, leaving from
  \texttt{[ORIGIN\_CITY]} and starting on or after \texttt{[START\_DATE]}. You are leaving with
  your hunting team and you will be a total of \texttt{[NUM\_ADULTS]} adults. You have a
  budget of \texttt{[PRICE\_MAX]}. You want to compare the packages between the different
  cities and book one, the one that will take you to your destiny.''
\end{example}
These templates were meant to add variety to the dialogues. The generic templates were also important for the users to create their own character and personality. We found that the combination of the two types of templates prevented the task from becoming too repetitive. Notably, we distributed the role-playing templates throughout the data collection process to bring some novelty and surprise. We also asked the participants to write templates (13 of them) to keep them engaged in the task.

To control data collection, we gave a set of instructions to the participants. The users received the following instructions:
\begin{itemize}[noitemsep]
\item Do not use uncommon slang terms, but feel free to use colloquialisms.
\item Make up personalities.
\item Feel free to end the conversation at any time.
\item Try to spell things correctly.
\item You do not necessarily have to choose an option.
\item Try to determine what you can get for your money.
\end{itemize}
These instructions were meant to encourage a variety of behaviours from the users. As for the wizards, they were asked to only talk about the database results and the task-at-hand. This is necessary if one wants to build a dialogue system that emulates the wizards' behaviour in this corpus. The wizard instructions were as follows:
\begin{itemize}[noitemsep]
\item Be polite, and do not jump in on the role play of the users.
\item Vary the way you answer the user, sometimes you can say something that would be right at another point in a dialogue.
\item Your knowledge of the world is limited by your database.
\item Try to spell things correctly.
\end{itemize}
We asked the wizards to sometimes act badly (second point in the list). It is interesting from a dialogue management point of view to have examples of bad behaviour and of how it impacts user satisfaction. At the end of each dialogue, the user was asked to provide a wizard cooperativity rating on a scale of 1 to 5. The wizard, on the other hand, was shown the user's task and was asked whether she thought the user had accomplished it.

\subsection{Database Search Interface} 
Wizards received a link to a search interface every time a user was connected with them. The search interface was a simple GUI with all the searchable fields in the database (see \cref{db_ov}). For every search in the database, up to 10 results were displayed. These results were sorted by increasing price.

\subsection{Suggestions}
When a database query returned no results, suggestions were sometimes displayed to the wizards. Suggestions were packages obtained by relaxing one or more constraints. It was up to the wizard to decide whether or not to use suggestions. Our goal with suggestions is not to learn a recommender system, but to learn the timing of recommendation, hence the randomness of the mechanism.

\section{Statistics of the Corpus}
\label{corpus_characteristics}
We used the data collection process described in the previous section to collect \emph{\ndialogues{} dialogues}. 
\begin{figure}
	\begin{center}
    \begin{subfigure}[b]{0.40\textwidth}
        \includegraphics[width=\textwidth]{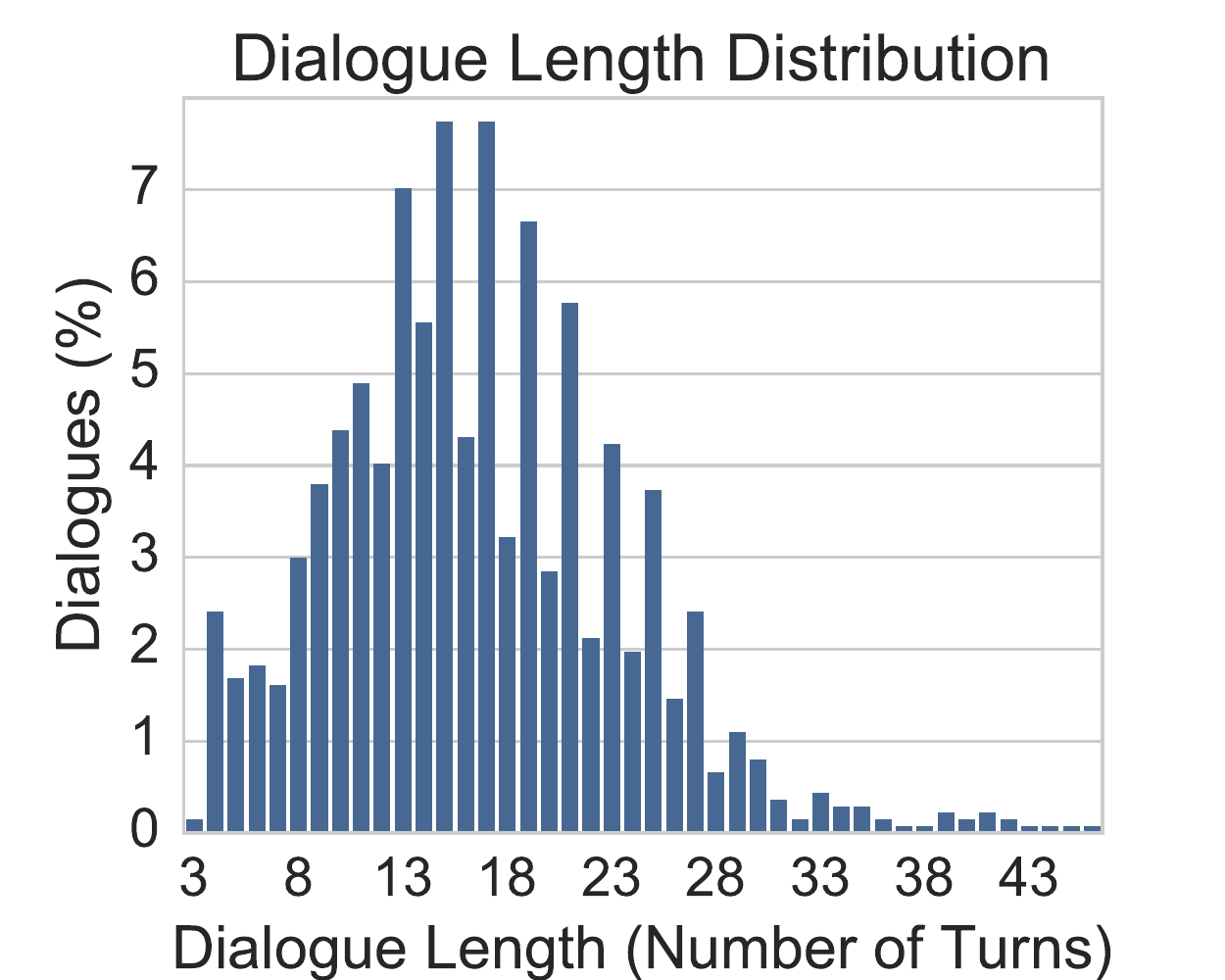}
        \caption{}
        \label{fig:turns}
        \vspace{0.2cm}
    \end{subfigure}
    \begin{subfigure}[b]{0.40\textwidth}
        \includegraphics[width=\textwidth]{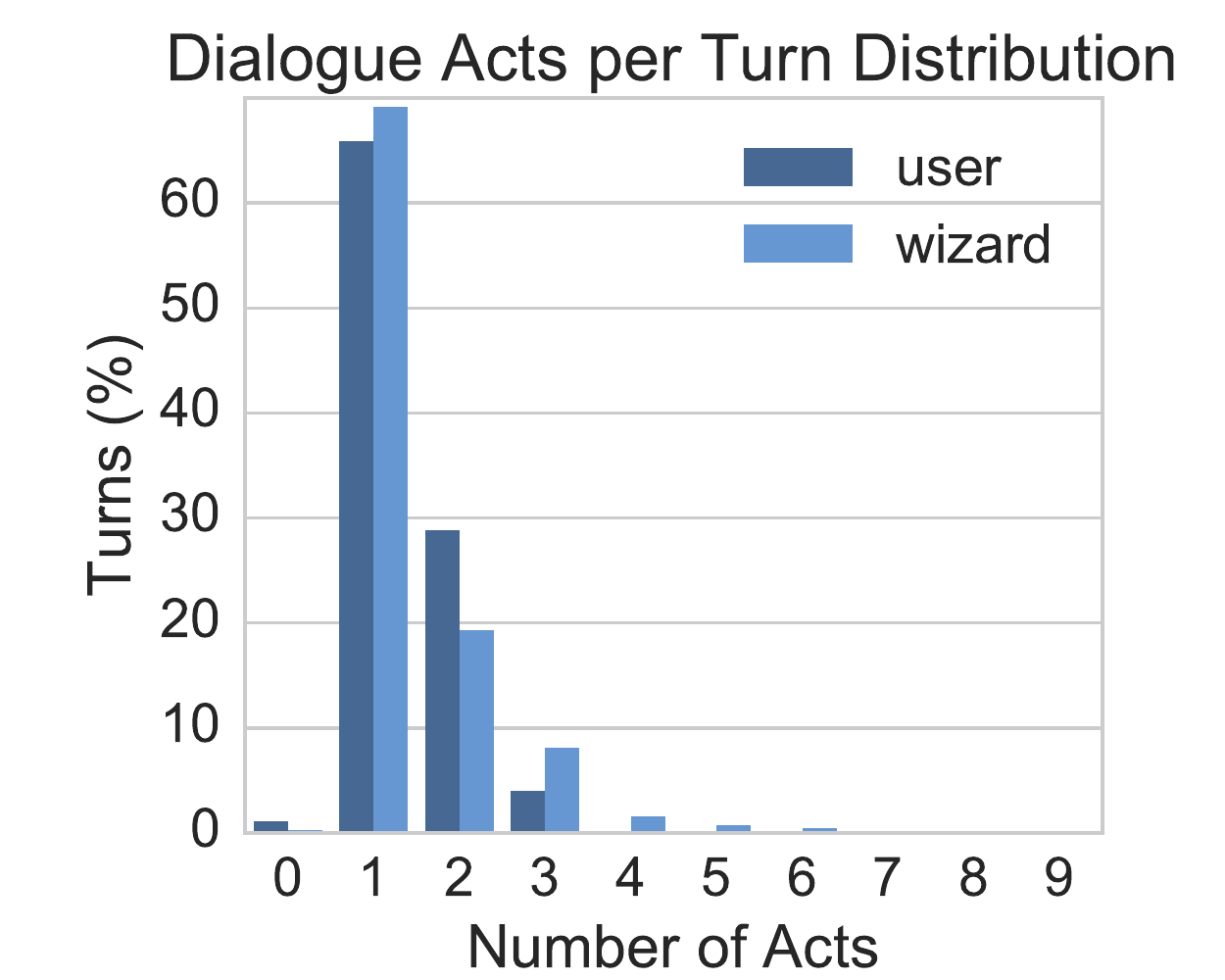}
        \caption{}
        \label{fig:nb_acts}
        \vspace{0.2cm}
    \end{subfigure}
    \begin{subfigure}[b]{0.40\textwidth}
        \includegraphics[width=\textwidth]{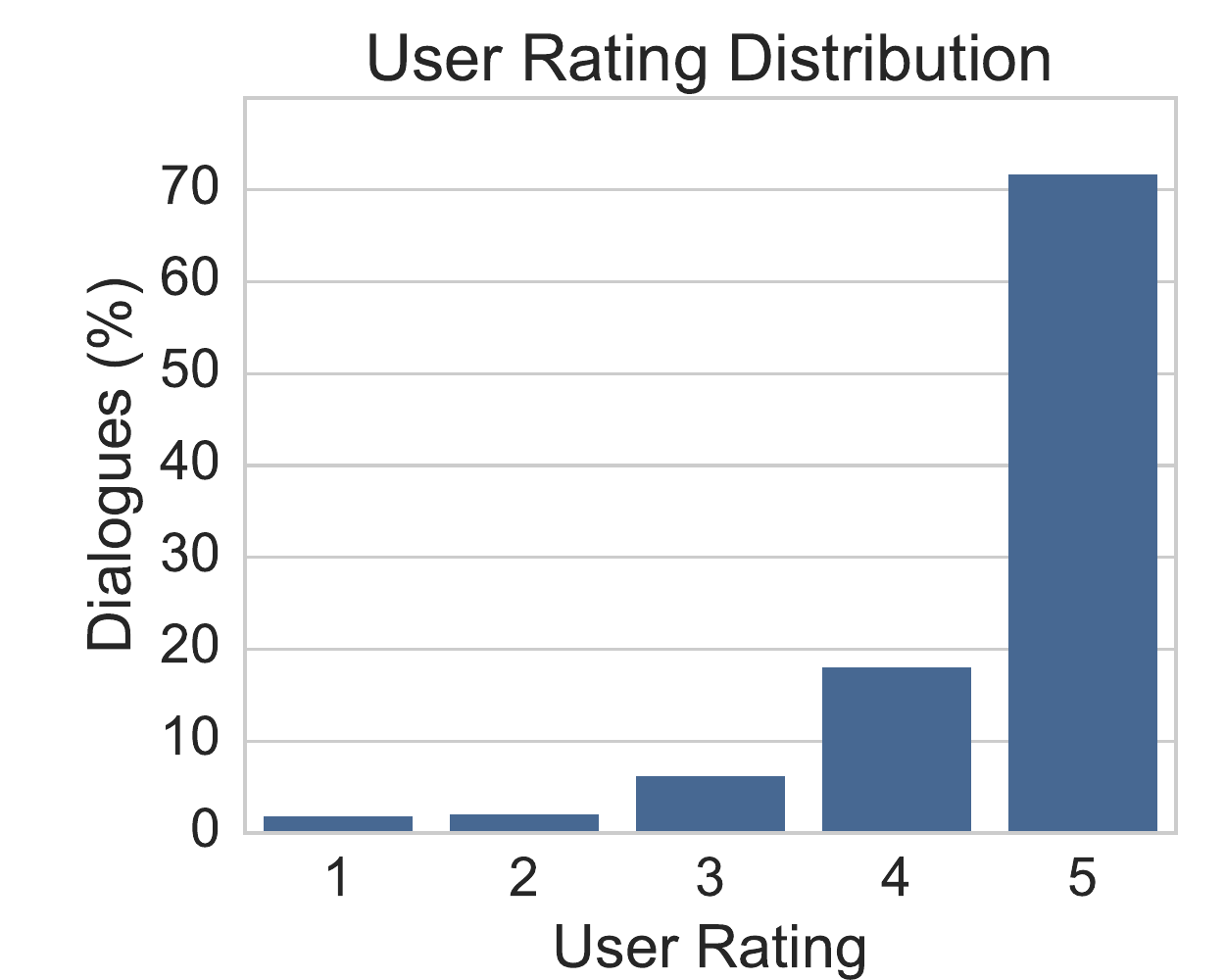}
        \caption{}
        \label{fig:u_ratings}
    \end{subfigure}
    \begin{subfigure}[b]{0.40\textwidth}
        \includegraphics[width=\textwidth]{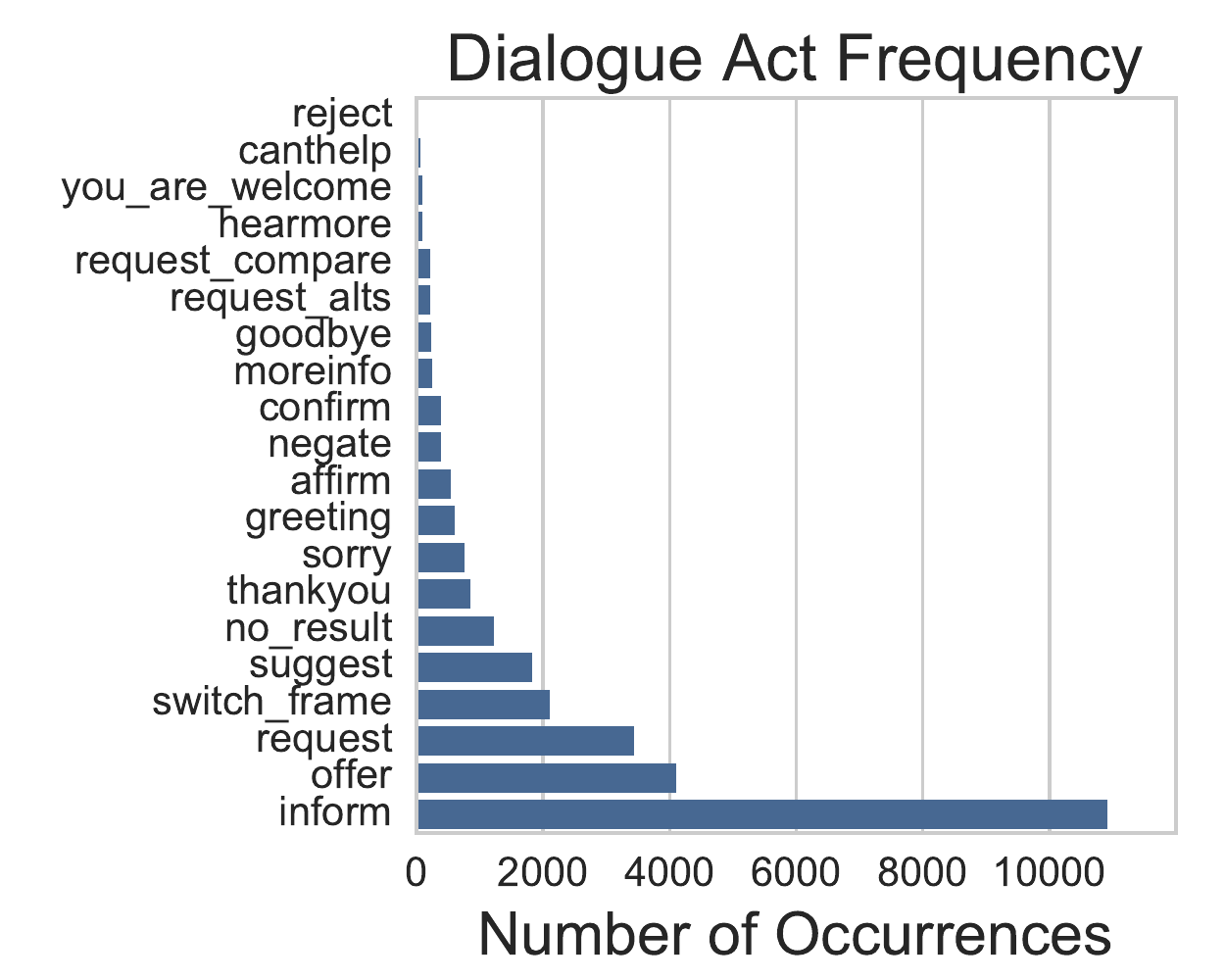}
        \caption{}
        \label{fig:act_occurrences}
    \end{subfigure}
    \caption{Overview of the Frames corpus}
    \label{fig:global_stats}
    \end{center}
\end{figure}
\Cref{fig:turns} shows the distribution of dialogue length in the corpus. The average number of turns is 15, for a total of \emph{19986 turns} in the dataset. A turn is defined as a Slack message sent by either a user or a wizard. A user turn is always followed by a wizard turn and vice versa.

\Cref{fig:nb_acts} shows the number of acts per dialogue turn. About 25\% of the dialogue turns have more than one dialogue act. The turns with 0 dialogue act are turns where the user asked for something that the wizard could not provide, \textit{e.g.},~because it was not part of the database. An example in the dataset is: ``Would my room have a view of the city? How much would it cost to upgrade to a room with a view?''. We left such user turns unannotated and they are usually followed up by the wizard saying she cannot provide the required information.

\Cref{fig:u_ratings} shows the distribution of user ratings. More than 70\% of the dialogues have the maximum rating of 5. \Cref{fig:act_occurrences} shows the occurrences of dialogue acts in the corpus. The dialogue acts are described in \cref{tab:dialogue_acts} in \cref{app:dialogue_acts}.  We present the annotation scheme in the following section.

\section{Dialogue Annotation Scheme}
\label{annotation}
We annotated the Frames dataset with three types of labels:
\begin{enumerate}
\item Dialogue acts, slot types, slot values, and references to other frames for each utterance.
\item The ID of the currently active frame.
\item Frame labels which were automatically computed based on the previous two sets of labels.
\end{enumerate}

\subsection{Dialogue Acts, Slot Types, and Slot Values}
\label{sec:dialogue-acts-slot}
Most of the dialogue acts used for annotation are acts which are usually encountered in the goal-oriented setting such as \texttt{inform} and \texttt{offer}.
We also introduced dialogue acts which are specific to our frame tracking setting such as \texttt{switch\_frame} and \texttt{request\_compare}. The dialogue acts are listed in \cref{tab:dialogue_acts}.

Our annotation uses three sets of slot types. The first set, listed in \cref{tab:searchable_fields,tab:displayed_fields}, corresponds to the fields of the database.
The second set is listed in \cref{tab:slot_type} and contains the slot types which we defined in order to describe specific aspects of the dialogue such as \texttt{intent} and \texttt{action}. An \texttt{intent} indicates whether or not the user wants to book a package, whereas an \texttt{action} indicates whether or not the wizard should, or did, book it.
We also introduced several \texttt{count} slot types which were used most often by wizards to summarize information in the database, \textit{e.g.},~``I have 2 hotels in Marseille''. In this case, the wizard informs that the count for hotels is 2.

The remaining slot types in \cref{tab:slot_type} were introduced to describe frames and cross-references between them.
Before we discuss these slot types, we define frames more formally in the following section.

\subsection{Frames}
\subsubsection{Definition}
\label{frameDef}
Semantic frames form the core of our dataset. A semantic frame is defined by the following four components:
\begin{itemize}[noitemsep]
\item User requests: slots whose values the user wants to know for this frame.
\item User binary questions: user questions with slot types and slot values.
\item Constraints: slots which have been set to a particular value by the user or the wizard.
\item User comparison requests: slots whose values the user wants to know for this frame and one or more other frames.
\end{itemize}
Several of these labels are used in the Dialogue State Tracking Challenge (DSTC) \citep{Williams:16b}. In DSTC, a semantic frame contains the constraints set by the user, the user requests, and the user's search method (\textit{e.g.}, by constraints or alternatives). In our case, constraints can also be set by the wizard when she suggests or offers a package. Any field in the database (see \cref{tab:searchable_fields,tab:displayed_fields} in \cref{db_ov}) can be constrained by the user or the wizard. The comparison requests and the binary questions were added after analysing the dialogues. The comparison requests correspond to the \texttt{request\_compare} dialogue act. This dialogue act is used to annotate turns when a user asks to compare different results, for instance: ``\textit{Could you tell me which of these resorts offers free wifi?}''. These questions relate to several frames. Binary questions are questions with slot types and slot values, \textit{e.g.},~``\textit{Is this hotel in the downtown area of the city?}'' (\texttt{request} act), or ``\textit{Is the trip to Marseille cheaper than to Naples?}'' (\texttt{request\_compare} act), as well as all \texttt{confirm} acts. Binary questions concern one or several frames.

\subsubsection{Creation and Annotation}
Each dialogue starts in frame~1. New frames are introduced when the wizard offers or suggests something, or when the user modifies pre-established slots. Thus, all values discussed during the dialogue are recorded and the user can return to a previous set of constraints at any point.
An example is given in \cref{tab:example_frames}. In this example, the frame number changes when the user modifies several slot values: the destination city, the number of adults for the trip, and the budget.
Though frames are created for each offer or suggestion made by the wizard, the \textit{active} frame can only be changed by the user so that the user has control over the dialogue.
If the user asks for more information about a specific offer or suggestion, the active frame is changed to the frame introduced with that offer or suggestion. This change of frame is indicated by a \texttt{switch\_frame} act (see \cref{db_ov}). 
The rules for creating and switching frames are summarized in \cref{tab:frames_creation}.
\begin{table}[h!]
\begin{center}
\caption{Dialogue excerpt with active frame annotation}
\begin{tabu}to\linewidth{@{}X[l,1]X[l,8]X[c,1]@{}} 
\toprule
\textbf{Author} & \textbf{Utterance}                                                            & \textbf{Frame} \\\midrule
User            & I'd like to book a trip to Atlantis from Caprica on Saturday,                    & 1              \\
                & August 13, 2016 for 8 adults. I have a tight budget of 1700.                  &                \\
Wizard          & \wizardcell{Hi...I checked a few options for you, and unfortunately,}                      & 1              \\
                & \wizardcell{we do not currently have any trips that meet this criteria.}                   &                \\
                & \wizardcell{Would you like to book an alternate travel option?}                            &                \\
User            & Yes, how about going to Neverland from Caprica on August 13,                     & 2              \\
                & 2016 for 5 adults. For this trip, my budget would be 1900.                    &                \\
Wizard          & \wizardcell{I checked the availability for those dates and there were no trips available.} & 2              \\
                & \wizardcell{Would you like to select some alternate dates?}                                &                \\
\bottomrule
\end{tabu}
\label{tab:example_frames}
\end{center}
\end{table}

\begin{table}
\begin{center}
\caption{Frequency of frame creation and switching events}
\begin{tabu}to\linewidth{@{}X[l,1.5]X[l,1]X[l,5]S[table-format=2]S[table-format=4]@{}} 
\toprule
\textbf{Rule Type} & \textbf{Author} & \textbf{Rule Description}                                                                              & \multicolumn{2}{c}{\textbf{Frequency}}    \\\cmidrule{4-5}
                   &                 &                                                                                                        & {\textbf{Relative}} & {\textbf{Absolute}} \\\cmidrule{1-5}
Creation           & User            & Changing the value of a slot                                                                           & \SI{31}{\percent}          & 2092                \\
                   & Wizard          & Making an offer or a suggestion                                                                        & \SI{69}{\percent}          & 4762                \\\midrule
Switching          & User            & Changing the value of a slot (it causes the dialogue to switch to that frame) & \SI{50}{\percent}                & 2092                \\
                   &                 & Considering a wizard offer or suggestion                                                               & \SI{39}{\percent}          & 1635                \\
                   &                 & Switching to an earlier frame by mentioning its slot values                                            & \SI{11}{\percent}          & 458                 \\
\bottomrule
\end{tabu}
\label{tab:frames_creation}
\end{center}
\end{table}

We introduced specific slot types for recording the creation and modification of frames. These slot types are \texttt{id}, \texttt{ref}, \texttt{read}, and \texttt{write} (see \cref{tab:slot_type} in \cref{app:dialogue_acts}). The frame id is defined when the frame is created and is used to switch to this frame when the user decides to do so.

The other slot types -- \texttt{ref}, \texttt{read}, and \texttt{write} -- are used to annotate cross-references between frames, which are a crucial component of the recorded dialogues. A reference has two parts: the number of the frame it is referring to and the slots and values that are used to refer to that frame (if any). For instance, \texttt{ref[1\{name=Tropic\}]} means that frame 1 is being referred to by the hotel name \textit{Tropic}.
If anaphora is used to refer to a frame, we annotated this with the slot \texttt{ref\_anaphora} (\textit{e.g.},~``This is too long'' -- \texttt{inform(duration=too long,ref\_anaphora=this)}).
Inside an \texttt{offer} dialogue act, a \texttt{ref} means that the frame corresponding to the offer is derived from another frame. For example, here is an utterance from the corpus, written by a wizard:
\begin{example}
``Here are a couple of options. The first option is a 3.0 star hotel (the Tropic), with a guest rating of 4.77/10 and a business class flight. The cost is 1002.27 USD. Or, if you prefer, you could choose the same 3.0 star hotel with a guest rating of 4.77/10 (the Tropic) and an economy flight, for 812.69.''
\end{example}
This utterance is annotated with the following dialogue acts:
\begin{itemize}[noitemsep]
\item \texttt{offer(category=3.0,name=Tropic,gst\_rating=4.77/10,id=6)};
\item\texttt{offer(ref=[6],seat=business,price=1002.27 USD,id=7)};
\item and \texttt{offer(ref=[6],seat=economy,price=812.69,id=8)}.
\end{itemize}
Here, the frames corresponding to the last two offers are derived from the first one by inheriting all values. 

The slot types \texttt{read} and \texttt{write} only occur inside a wizard's \texttt{inform} act and are used by the wizards to provide relations between offers or suggestions: \texttt{read} is used to indicate which frame the values are coming from (and which slots are used to refer to this frame, if any), while \texttt{write} indicates the frame where the slot values are to be written (and which slot values are used to refer to this frame, if any). If there is a \texttt{read} without a \texttt{write}, the current frame is assumed as the storage for the slot values. A slot type without a value indicates that the value is the same as in the referenced frame, but was not mentioned explicitly \textit{i.e.}, ``for the same price''.

\Cref{tab:example_read_write} gives an example of how these slot types are used in practice: \texttt{inform(\linebreak[1]\emph{read}=[7\{dst\_city=Punta Cana, category=2.5\}]} means that the values \textit{2.5} and \textit{Punta Cana} are to be read from frame~7, and to be written in the current frame. At this turn of the dialogue, the wizard repeats information from frame~7.

The annotation  \texttt{inform(breakfast=False, \emph{write}=[7\{name=El Mar\}])} means that the value \textit{False} for breakfast is written in frame~7 and that frame~7 was identified in this utterance by the name of the hotel \textit{El Mar}. 

\begin{table}[tbp]
\begin{center}
\caption{Annotation example with the \texttt{write} and \texttt{read} slot types}
\begin{tabu}to\linewidth{@{}X[l,1] X[l,4.7] X[c,0.8] X[l,4]@{}} 
\toprule
\textbf{Author} & \textbf{Utterance}                       & \textbf{Frame} & \textbf{Annotation}                      \\\midrule
Wizard          & \wizardcell{I am only able to find hotels with a}     & 6              & inform(\emph{read}=[7\{dst\_city=Punta Cana, \\
                & \wizardcell{2.5 star rating in Punta Cana for that time.} &                & \hspace*{2mm}category=2.5\}])                         \\
User            &	2.5 stars will do.                       & 11             & inform(category=2.5)                     \\
                & Can you offer any additional activities? &                &                                          \\
Wizard          & \wizardcell{Unfortunately I am not able to provide}   & 11             & sorry, canthelp                          \\
                & \wizardcell{this information.}                        &                &                                          \\
User            & How about breakfast?                     & 11             & request(breakfast)                       \\
Wizard          & \wizardcell{El Mar does not provide breakfast.}       & 11             & inform(breakfast=False,                      \\
                &                                          &                & \hspace*{2mm}\emph{write}=[7\{name=El Mar\}])         \\
\bottomrule
\end{tabu}
\label{tab:example_read_write}
\end{center}
\end{table}

\begin{figure}[b!]
	\begin{center}
	\includegraphics[scale=0.6]{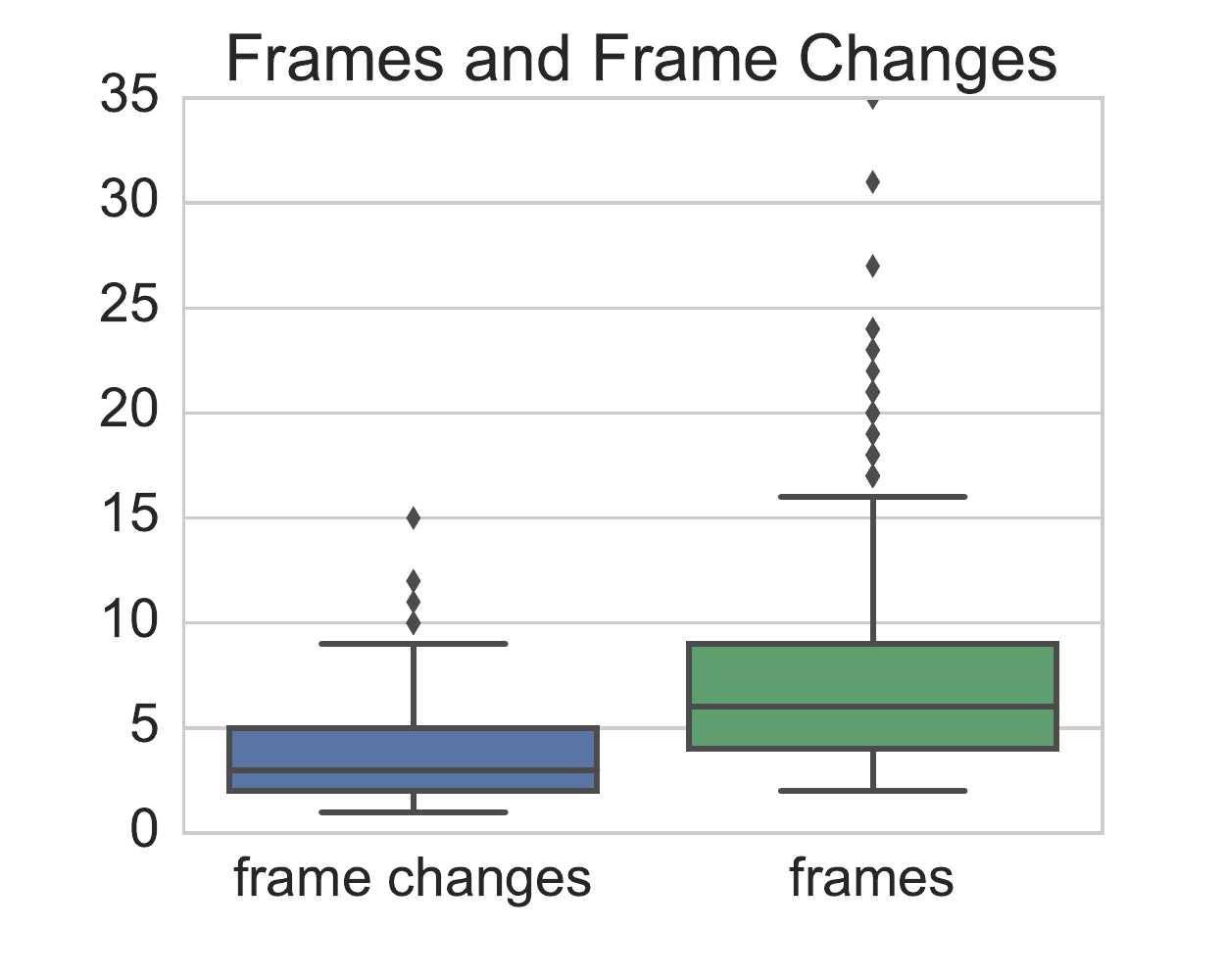}
    \caption{Number of frames and frame switches in the corpus}
    \label{fig:frames_stats}
    \end{center}
\end{figure}

The average number of frames created per dialogue is 6.71 and the average number of frame switches is 3.58. \Cref{fig:frames_stats} shows boxplots for the number of frame creations and the number of frame changes in the corpus.

\section{Research Topics}
\label{sec:challenges}
Frames can be used to research many aspects of goal-oriented dialogue, from Natural Language Understanding (NLU) to natural language generation. In this section, we propose three topics that we believe are new and representative of this dataset.
\subsection{Frame Tracking}
\subsubsection{Definition}
\label{frameTrackDef}
\emph{Frame tracking} extends state tracking \citep{Henderson:15,Williams:16b} to a setting where several semantic frames are tracked simultaneously. In state tracking, the dialogue history is compressed into one semantic frame. Essentially, this implies that every new slot value overwrites the previous one, which prevents the user from comparing options or returning to an item discussed earlier. In frame tracking, a new value creates a new semantic frame. The frame tracking task is significantly harder as it requires, for each user utterance, identifying the active frame as well as all the frames modified by the utterance.

\begin{definition}[Frame Tracking]
At each user turn $t$, we assume access to the full dialogue history $H = \{f_1,...,f_{n_{t-1}}\}$, where $f_i$ is a frame and $n_{t-1}$ is the number of frames created so far in the dialogue. For a user utterance $u_t$ at time $t$, we provide the following NLU labels: dialogue acts, slot types, and slot values. The goal of frame tracking is to predict if a new frame is created and to predict for each dialogue act the \texttt{ref} labels (possibly none) and the \texttt{id}s of the frames referenced.
\end{definition}

Predicting the frame that is referenced by a dialogue act requires detecting if a new frame is created and recognizing a previous frame from the values mentioned by the user (potentially a synonym, \textit{e.g.}, NYC for New York), or by using the user utterance directly. It is necessary in many cases to use the user utterance directly because users do not always use slot values to refer to previous frames. An example in the corpus is a user asking: ``Which package has the soonest departure?''. In this case, the user refers to several frames (the packages) without ever explicitly describing which ones. This phenomenon is quite common for dialogue acts such as \texttt{switch\_frame} (979 occurrences in the corpus) and \texttt{request\_compare} (455 occurrences in the corpus). These cases can only be resolved by working on the text directly and solving anaphora.

\subsubsection{Evaluation Metrics}
\label{frameTrackEval}
We define two metrics: frame identification and frame creation. For frame identification, for each dialogue act, we compare the ground truth pair (\texttt{key-value}, \texttt{frame}) to that predicted by the frame tracker. We compute performance as the number of correct predictions over the number of pairs. A prediction is correct if and only if the \texttt{frame}, \texttt{key}, and \texttt{value} are the same in the ground truth and prediction. The \texttt{frame} is the \texttt{id} of the referenced frame. The \texttt{key} and \texttt{value} are respectively the type and the value of the slot used to refer to the frame (these can be null). 

For frame creation, we compute the number of times the frame tracker correctly predicts that a frame is created or correctly predicts that a frame has not been created over the number of dialogue turns.

\subsubsection{Related Work}
Frame tracking is closely related to state tracking in that it extends the task from only tracking the current user goal to tracking all the user goals that occur during the dialogue.

Recent approaches to state tracking have been suggested to go beyond the sequential slot-filling approach. An important contribution is the Task Lineage-based Dialog State Tracking (TL-DST) proposed by \citep{Lee:16}. TL-DST is a framework that allows keeping track of tasks across different domains. Similarly to frame tracking, \citeauthor{Lee:16} propose building a dynamic structure of the dialogue containing different frames corresponding to different tasks. They defined different sub-tasks among which \textit{task frame parsing} which is closely related to frame tracking except that they impose constraints on how a dialogue act can be assigned to a frame and a dialogue act can only relate to one frame. Because of the lack of data, \citet{Lee:16} trained their tracking model on datasets released for DSTC (DSTC2 and DSTC3, \citealp{Henderson:14,Henderson:14c}). As a result, they could artificially mix different tasks, \textit{e.g.}, looking for a restaurant and looking for a pub, but they could not study how human beings switch between topics. In addition, this framework can switch between different tasks but does not handle comparisons which is an important aspect of frame tracking. 

Another related approach was proposed by \citep{Perez:16} who framed the state tracking task as a question-answering task. Their state tracker is based on a memory network \citep{weston:14} and can answer questions about the user goal at the end of the dialogue. They also propose adding functionalities such as keeping a list of the constraints expressed by the user during the dialogue. 

We propose the dataset in order to encourage more research on complex state tracking behaviours. In addition, we propose the frame tracking task as a principled way of modelling such behaviour in the case of decision-making but researchers are free to use this dataset for any task that they define.

\subsection{Dialogue Management}
One of the notable aspects of this dataset is that memory is not only a matter of frame tracking. Most of the time, the wizard would speak about the current frame to ask or answer questions. However, sometimes, the wizard would talk about previous frames. We can see it as appealing to memories in a conversation. An example is given in \cref{tab:dialogue_memory_wiz}. In the bold utterance in this dialogue, even though the active frame is frame 4, the wizard mentions a previous frame (frame 3). In order to reproduce this kind of behaviour, a dialogue manager would need to be able to identify potentially relevant frames for the current turn and to output actions for these frames.

\begin{table}[tbp]
\begin{center}
\caption{Dialogue excerpt where the wizard talks about a frame other than the active frame.}
  \begin{tabu}to\linewidth{@{}X[l,1]X[8,l]X[c,1]} 
\toprule
\textbf{Author} & \textbf{Utterance}                                                                    & \textbf{Frame} \\\midrule
User            & i need a vacation                                                                     & 1              \\
Wizard          & \wizardcell{How can I help?}                                                          & 1              \\ 
User            & i've got a few days off from august 26-august 31.                                     & 1              \\
                & im not flexible on this, but i still want to somehow treat myself                     &                \\  
                & with an 8 day trip (??) im leaving Dallas and i wanna check out mannheim                 &                \\  
Wizard          & \wizardcell{would a 5 day trip suffice?}                                              & 1              \\ 
User            & sure dude                                                                             & 2              \\ 
Wizard          & \wizardcell{A 5 star hotel called the Regal Resort, it has free wifi and a spa.}      & 2              \\ 
User            & dates?                                                                                & 3              \\ 
Wizard          & \wizardcell{Starts on august 27th until the 30th}                                     & 3              \\ 
User            & ok that could work.                                                                   & 4              \\ 
                & I would like to see my options in Santos as well                                    &                \\ 
\textit{Wizard} & \wizardcell{\textit{regal resort goes for \$2800 or there is the Hotel Globetrotter}} & \textit{4}     \\
                & \wizardcell{\textit{in Santos it has 3 stars and comes with breakfast and wifi,}}   &                \\
                & \wizardcell{\textit{it leaves on the 25th and returns on the 30th! all for \$2000}}   &                \\ 
User            & ahh I can't leave until august 26 though                                              & 4              \\ 
\textbf{Wizard} & \wizardcell{\textbf{then i guess you might have to choose the Regal resort}}                       & \textbf{4}     \\ 
User            & yeah. I will book it                                                                  & 3              \\ 
Wizard          & \wizardcell{Thank you!}                                                               & 3              \\
\bottomrule
\end{tabu}
\label{tab:dialogue_memory_wiz}
\end{center}
\end{table}

\Cref{tab:dialogue_memory_wiz} also illustrates another novelty. In the utterance in italic, the wizard actually performs two actions. The first action consists of informing the user about the price of the \textit{regal resort} and the second action consists of proposing another option, \textit{Hotel Globetrotter}. Performing more than one action per turn is a challenge when using reinforcement learning \citep{Pietquin:11,Gasic:12,Fatemi:16} and, to our knowledge, this has only been done in a simulated setting \citep{Laroche:09}.

\subsection{Natural Language Generation}
An interesting behaviour observed in our dataset is that wizards often tended to summarize database results. An example is the wizard saying: \textit{``The cheapest available flight is 1947.14USD.''} In this case, the wizard informs the user that the database has no cheaper result than the one she is proposing. To imitate this behaviour, a dialogue system would need to reason over the database and decide how to present the results to the user.

\section{Dataset Format}
\label{dataset_format}
\subsection{Dialogues}
We provide the Frames dialogues in JSON format. Each dialogue has five main fields: \texttt{turns}, \texttt{labels}, \texttt{user\_id}, \texttt{wizard\_id}, and \texttt{id}. The ids are unique for each dialogue (\texttt{id}), each user (\texttt{user\_id}), and each wizard (\texttt{wizard\_id}). 

The \texttt{labels} have two fields: 
\begin{description}[noitemsep,font=$\bullet$ \ttfamily\mdseries,style=unboxed]
\item [userSurveyRating] user rating of wizard cooperativity on a scale of 1 to 5 (see Section \ref{sec:data_collection}).
\item [wizardSurveyTaskSuccessful] wizard's perceived task completion (see Section \ref{sec:data_collection}). 
\end{description}

The \texttt{turns} have the following fields:
\begin{description}[noitemsep,font=$\bullet$ \ttfamily\mdseries,itemindent=-1.5mm]
\item [author] ``user'' or ``wizard''.
\item [text] the author's utterance.
\item [labels] the id of the currently active frame (\texttt{active\_frame}) as well as a list of dialogue acts (\texttt{acts}) each with a \texttt{name}, and \texttt{args} (key-value pairs), and a list of dialogue acts without \texttt{ref} tags (\texttt{acts\_without\_refs}) for frame tracking.
\item [timestamp] timestamp for the message.
\item [frames] List of all frames after integrating the current turn. Each frame has the following labels:
\begin{description}[noitemsep,font=$\bullet$ \ttfamily\mdseries,itemindent=-1.5mm]
\item [frame\_id] id of the frame.
\item [frame\_parent\_id] id of the parent frame.
\item [requests,] \texttt{binary\_questions, compare\_requests} user questions (see \cref{frameDef}). 
\item [info] properties of the frame (see \cref{frameDef}) Note that each slot can have multiple values, which accumulate as long as the frame does not change. For example, price can be both ``1000 USD'' and ``cheapest''.
  Each value has a boolean property ``negated'', expressing whether the user negated the value of the corresponding slot, for instance \textit{``I don't want to stay 3 days''} (\texttt{negate(duration=3)}), or negated an explicit confirmation question.
  When a user switches to a frame, we assume the user accepts all information provided by the wizard for that frame as ``constraints''. We drop these additional constraints when a constraint is modified by the user, or the user requests alternatives. Our motivation for this scheme is to make frames more distinguishable and encourage methods which correctly identify frame switches.
  Additionally to slots and their values, we added the following fields to keep track of specific aspects of the dialogue: 
\begin{description}[font=$\bullet$ \ttfamily\mdseries,style=unboxed]
  \item [REJECTED] a boolean value expressing if the user negated or affirmed an offer made by the wizard (corresponds to a \texttt{negate} act that does not follow a question).
  \item [MOREINFO] a boolean value expressing whether the user wants to know more about this frame, which happens if the wizard withholds detail information (see \texttt{moreinfo} act).
\end{description}
\end{description}
\item [db] (wizard turns only) list of search queries made by the wizard with the associated search results/suggestions.
\end{description}

\subsection{Hotels}
\begin{figure}[!h]
	\begin{center}
    \begin{subfigure}[b]{0.47\textwidth}
        \includegraphics[width=\textwidth]{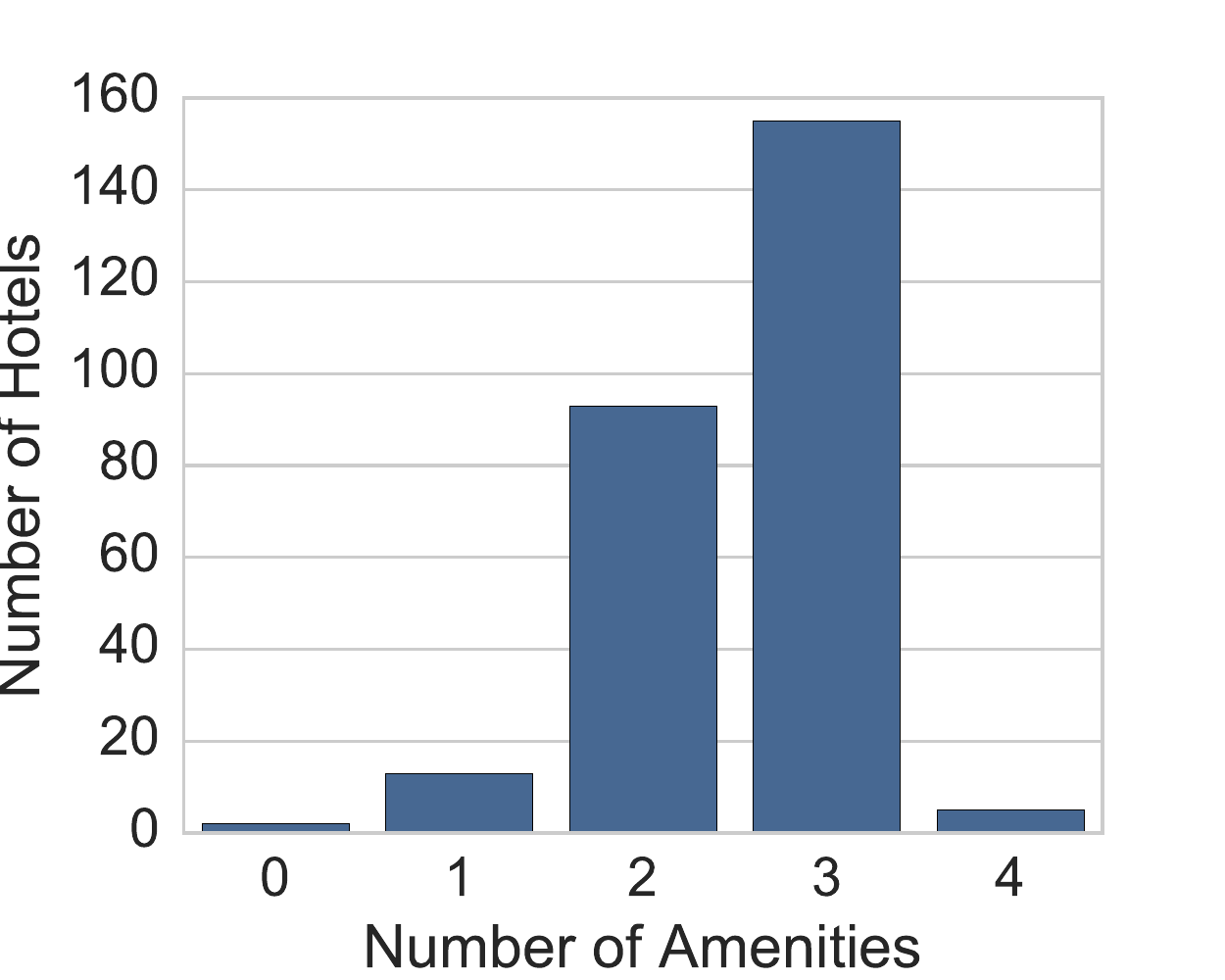}
        \label{fig:amenities}
        \vspace{0.2cm}
    \end{subfigure}
    \begin{subfigure}[b]{0.47\textwidth}
        \includegraphics[width=\textwidth]{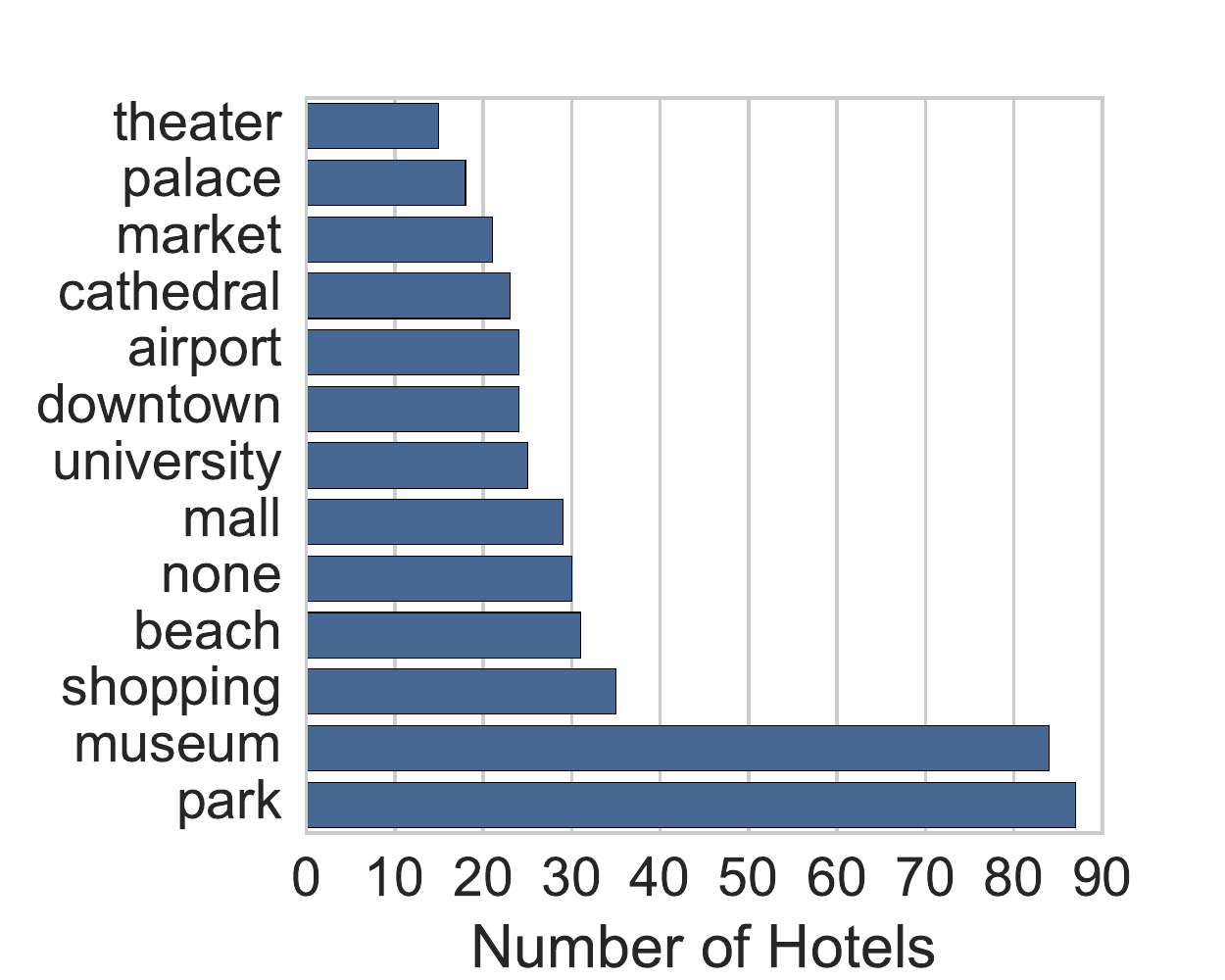}
        \label{fig:vicinity}
        \vspace{0.2cm}
    \end{subfigure}
    \caption{Left -- number of amenities per hotel. Right -- number of hotels in the vicinity of points of interest (none if the hotel is not close to any point of interest).}
    \label{fig:hotels_stats}
    \end{center}
\end{figure}

The vacation packages were generated randomly. A database of packages can be created by using the search results in the JSON files containing the dialogues. The hotels in these search results have all the fields listed in the \textit{Hotel Properties} section of \cref{tab:displayed_fields} in \cref{db_ov}. Note that amenities or points of interest in the vicinity of the hotel are only listed in a hotel's description if they are true. For instance, the field \textit{breakfast} is only present for hotels proposing free breakfast. \Cref{fig:hotels_stats} shows statistics for these boolean values.

\section{Baselines}
\subsection{Natural Language Understanding}
\label{baseline}
\begin{figure}[!h]
\centering
\includegraphics[scale=1]{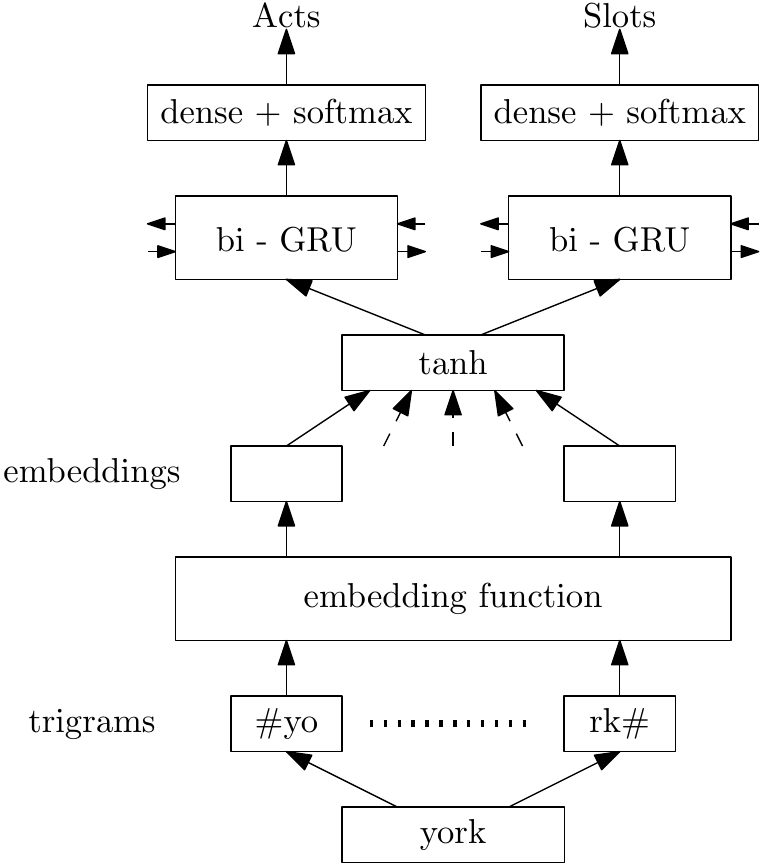}
\caption{Illustration of the NLU model for slots and acts prediction, taking input words and outputting labels for slots and acts. The model splits into slots-specific and acts-specific predictors after the word embedding layer, which computes a non-linearity on top of the per-word sum of character trigram embeddings.
\label{fig:baseline}}
\end{figure}
%

We define the NLU task as dialogue act prediction and IOB (Inside, Outside, Beginning) tagging. The IOB tag format (Inside, Outside, Beginning) is a common word-level annotation format for natural language understanding. A word tagged with \emph{O} means that this word is not part of any slot value. The \emph{B} and \emph{I} tags are used for slot values. Every word which is the beginning of a slot value is tagged with \emph{B} and the \emph{I} tag is used for subsequent words until the end of the slot value. For instance, ``I need to go to New York for business'' would be tagged as ``I (O) need (O) to (O) go (O) to (O) New (B) York (I) for (O) business (O)''. We generated these word-level tags by matching the slot values in the manual annotations with the corresponding textual utterances. The act tags were also generated at the word level: for a given dialogue act with slot values, each word between the slot value that occurred first in the text and the one that occurred last in the text was tagged with the corresponding act. The other words were tagged with \emph{O}.

The NLU model is illustrated in \cref{fig:baseline}. The IOB tagging part operates on character trigrams and is based on the robust named entity recognition model \citep{Arnold:16}. We predict, for each word of the utterance, a pair of tags -- one for the act and one for the slot. The model splits into two parts: one part is trained to predict dialogue acts and 
the other part is trained to predict slot types (at this stage, we predict either a slot type or an \emph{O} tag). These two parts share an embedding matrix for the input character trigrams. Note that the model only predicts IOB tags for slots whose values can be found in the text. Therefore, the prediction for slots such as \texttt{intent} or vicinities and amenities is not evaluated for this simple baseline.

The two parts of the model are trained simultaneously, using a modified categorical crossentropy loss for either set of outputs. We modify the loss to ignore \emph{O} labels that are already predicted correctly by the model. We introduce this modification because \emph{O} labels are far more frequent than other labels, and not limiting their contribution to the loss causes the model to get stuck into a mode where it predicts \emph{O} labels for every word. The loss for the two parts of the model are added together, and the combined objective is optimized using the ADAM optimizer \citep{Kingma:14}. 

We provide F1 scores for acts and slots for this model in Table \ref{tab:nlu_results}. We report average and standard deviation over ten leave-one-user-out splits of the Frames dataset. We had a total of 11 participants in the user role during data collection. Two participants performed significantly fewer dialogues than the others. We merged the dialogues generated by these two participants (ids U21E41CQP and U23KPC9QV). For each of the resulting 10 users, we randomly split the combined dialogues of the nine others into training (80\%) and validation (20\%), and then tested on the dialogues from the held-out user.

\begin{table}
\begin{center}
\caption{F1 scores for the NLU baseline (mean and standard deviation).}
\begin{tabu}to\linewidth{@{}X[c]X[c]@{}}
\toprule
                   {Dialogue Acts} & {Slots}\\\midrule
                   0.78 $\pm$ 0.05 & 0.79 $\pm$ 0.04 \\
\bottomrule
\end{tabu}
\label{tab:nlu_results}
\end{center}
\end{table}

\begin{table*}[!t]
\begin{center}
\caption{Performance of the Frame Tracking Baselines (mean and standard deviation).}
\begin{tabu}to\linewidth{@{}X[l]X[1.6,l]X[1.5,l]@{}}
\toprule
                   & {Rule-Based} & {Random}\\\midrule
Frame Creation  &   0.49 $\pm$ 0.03     &   0.47 $\pm$ 0.02 \\
Frame Identification & 0.24 $\pm$ 0.02  &  0.18 $\pm$ 0.02 \\
\bottomrule
\end{tabu}
\label{tab:baseline_results}
\end{center}
\end{table*}

\subsection{Frame Tracking}
The rule-based frame tracker takes as input the \texttt{acts\_without\_refs} annotation and the values set in the existing frames. We write $f[k]$ to denote the value of slot $k$ in frame $f$. According to hand-designed rules, the frame tracker predicts the \texttt{ref} tags (for frame identification, see \cref{frameTrackEval}) and frame creations. For an act $a(k{=}v)$ in frame $f$, the following rules are used:
\begin{description}[noitemsep,font=$\bullet$ \itshape\mdseries,leftmargin=3mm]
	\item [Create and switch to a new frame] if $a$ is \texttt{inform} and $f[k]$ is set, but $v$ does not match $f[k]$.
  \item [Switch to frame $g$] if $a$ is \texttt{switch\_frame} and $g[k]$ matches $v$. If no match is found, switch to the most recently created frame.\footnote{This is a reasonable assumption since this case often happens when a wizard makes an offer and the user talks about this offer}
  \item [Assign \texttt{ref} to frame $g$] if $a$ can have a \texttt{ref} tag, and $g[k]$ matches $v$. The most recently created frame is used in ambiguous cases. If no match is found, assign \texttt{ref} to the current frame.
\end{description}

We compare this baseline to random performance. For random performance, for each (dialogue act, slot type) combination, we computed priors on the corpus for each time the user would refer to the current frame \textit{vs} a previous one. We sampled whether each slot was referring to the current frame or another one based on that prior, and if it referred to another frame, the frame number for that other frame was sampled uniformly from the list of frames created so far.

\Cref{tab:baseline_results} presents results for these baselines. We report results over 10 runs following the same method as for the NLU model. \Cref{tab:baseline_results} shows that the rule-based baseline only performs slightly better than random on frame identification and performs similarly on frame creation. In general, these results suggest that simple rules are far from adequate for frame tracking and require more in-depth analysis of the user text.

\section{Conclusion and Future Work}
\label{conclusion}
In this paper, we introduced the Frames dataset: a corpus of human-human dialogues for researching the role of memory in goal-oriented dialogue systems. We propose this dataset to study memory in goal-oriented dialogue systems. We formalized the frame tracking task, which extends the state tracking task to a setting where several semantic frames are simultaneously tracked throughout the dialogue. We proposed a baseline for this task and we showed that there is a lot of room for improvement. Finally, we showed that Frames can be used to research other interesting aspects of dialogue such as the use of memory for dialogue management and information presentation through natural language generation. We propose adding memory as a first milestone towards goal-oriented dialogue systems that support more complex dialogue flows. Future work will consist of proposing models for frame tracking as well as proposing a methodology to scale up data collection and annotation.

{
  \setlength\bibitemsep{6pt}
  \printbibliography
}

\newpage

\appendix

\section{Database Overview}
\label{db_ov}

\begin{table}[!h]
\begin{center}
  \tabulinesep=2pt
\caption{Searchable fields in the database of packages}
\begin{tabu}to\linewidth{@{}>{\ttfamily}X[l,1]X[l,2.3]@{}} 
\toprule
\multicolumn{1}{@{}l}{\textbf{Field}}       & \textbf{Description}                                                   \\\midrule
PRICE\_MAX           & Maximum price the user is willing to pay                               \\
PRICE\_MIN           & Minimum price defined by the user                                      \\
DESTINATION\_CITY    & Destination city                                                       \\
MAX\_DURATION        & Maximum number of days for the trip                                    \\
NUM\_ADULTS          & Number of adults                                                       \\
NUM\_CHILDREN        & Number of children                                                     \\
START\_DATE          & Start date for the trip                                                \\
END\_DATE            & End date for the trip                                                  \\
ARE\_DATES\_FLEXIBLE & Boolean value indicating whether or not the user's dates are flexible. 
                     If True, then the search is broadened to 2 days before \texttt{START\_DATE} 
                      and 2 days after \texttt{END\_DATE}.                                            \\
ORIGIN\_CITY         & Origin city                                                            \\
\bottomrule
\end{tabu}
\label{tab:searchable_fields}
\end{center}
\end{table}

\begin{table}[!h]
\begin{center}
\tabulinesep=1.5pt
\caption{Non-searchable fields in the database of packages}
\begin{tabu}to\linewidth{@{}>{\ttfamily}X[1.1,l]X[1.0,l]@{}}
\toprule
\multicolumn{1}{@{}l}{\textbf{Field}}                                                                          & \textbf{Description}                                                                  \\\midrule
\multicolumn{1}{@{}l}{\textbf{Global Properties}}                                                              &                                                                                       \\ 
\cmidrule(r){1-1}
PRICE                                                                                   & Price of the trip including flights and hotel                                         \\ DURATION & Duration of the trip                              \\
\cmidrule(r){1-1} \multicolumn{1}{@{}l}{\textbf{Hotel Properties}}                                             &                                                                                       \\
\cmidrule(r){1-1}
NAME                                                                                    & Name of the hotel                                                                     \\
COUNTRY                                                                                 & Country where the hotel is located                                                    \\
CATEGORY                                                                                & Rating of the hotel (in number of stars)                                              \\
CITY                                                                                    & City where the hotel is located                                                       \\
GUEST\_RATING                                                                           & Rating of the hotel by guests (in number of stars)                                     \\
BREAKFAST,\,PARKING,\,WIFI,\,GYM,\,SPA & Boolean value indicating whether or not the hotel offers this amenity. \\
PARK,\,MUSEUM,\,BEACH,\,SHOPPING,\linebreak\hspace*{1mm}MARKET,\,AIRPORT,\,UNIVERSITY,\,MALL,\linebreak\hspace*{1mm}CATHEDRAL,\,DOWNTOWN,\,PALACE,\,THEATRE & Boolean value indicating whether or not the hotel is in the vicinity of one of these. \\
\cmidrule(r){1-1} \multicolumn{1}{@{}l}{\textbf{Flights Properties}}                                           &                                                                                       \\ \cmidrule(r){1-1}
SEAT                                                                                    & Seat type (economy or business)                                                       \\
DEPARTURE\_DATE\_DEP                                                                    & Date of departure to destination                                                      \\
DEPARTURE\_DATE\_ARR                                                                    & Date of return flight                                                                 \\
DEPARTURE\_TIME\_DEP                                                                    & Time of departure to destination                                                      \\
ARRIVAL\_TIME\_DEP                                                                      & Time of arrival to destination                                                        \\
DEPARTURE\_TIME\_ARR                                                                    & Time of departure from destination                                                    \\
ARRIVAL\_TIME\_ARR                                                                      & Time of arrival to origin city                                                        \\
DURATION\_DEP                                                                           & Duration of flight to destination                                                     \\
DURATION\_ARR                                                                           & Duration of return flight                                                             \\
\bottomrule
\end{tabu}
\label{tab:displayed_fields}
\end{center}
\end{table}

\newpage

\section{Dialogue Acts and Slot Types}
\label{app:dialogue_acts}

\begin{table}[!h]
\begin{center}
\caption{List of dialogue acts in the annotation of Frames}
\begin{tabu}to \linewidth{@{}>{\ttfamily}X[1,l]X[1,c]X[4,l]@{}} 
\toprule
\multicolumn{1}{@{}l}{\textbf{Dialogue Act}} & \textbf{Speaker} & \textbf{Description}                                          \\\midrule
inform                & User/Wizard      & Inform a slot value                                           \\
offer                 & Wizard           & Offer a package to the user                                   \\
request               & User/Wizard      & Ask for the value of a particular slot                        \\
switch\_frame         & User             & Switch to a frame                                             \\
suggest               & Wizard           & Suggest a slot value or package                               \\
                      &                  & that does not match the user's constraints                    \\
no\_result            & Wizard           & Tell the user that the database returned no results           \\
thankyou              & User/Wizard      & Thank the other speaker                                       \\
sorry                 & Wizard           & Apologize to the user                                         \\
greeting              & User/Wizard      & Greet the other speaker                                       \\
affirm                & User/Wizard      & Affirm something said by the other speaker                    \\
negate                & User/Wizard      & Negate something said by the other speaker                    \\
confirm               & User/Wizard      & Ask the other speaker to confirm a given slot value           \\
moreinfo              & User             & Ask for more information on a given set of results            \\
goodbye               & User/Wizard      & Say goodbye to the other speaker                              \\
request\_alts         & User             & Ask for other possibilities                                   \\
request\_compare      & User             & Ask the wizard to compare packages                            \\
hearmore              & Wizard           & Ask the user if she'd like to hear more about a given package \\
you\_are\_welcome     & Wizard           & Tell the user she is welcome                                  \\
canthelp              & Wizard           & Tell the user you cannot answer her request                   \\
reject                & Wizard           & Tell the user you did not understand what she meant           \\
\bottomrule
\end{tabu}
\label{tab:dialogue_acts}
\end{center}
\end{table}

\begin{table}[!h]
\begin{center}
  \tabulinesep=2pt
\caption{List of slot types not present in the database}
\begin{tabu}to\linewidth{@{}>{\ttfamily}X[l,1]X[l,2.7]@{}} 
\toprule
\multicolumn{1}{@{}l}{\textbf{Slot Type}} & \textbf{Description}                                                              \\ \midrule
count              & Number of different packages                                                      \\
count\_amenities   & Number of amenities                                                               \\
count\_name        & Number of different hotels                                                        \\
count\_dst\_city   & Number of destination cities                                                      \\
count\_seat        & Number of seat options (for flights)                                              \\
count\_category    & Number of star ratings                                                            \\
id                 & Id of the frame created (for offers and suggestions)                              \\ 
vicinity           & Vicinity of the hotel                                                             \\
amenities          & Amenities of the hotel                                                            \\
ref\_anaphora      & Words used to refer to a frame                                                    
                   \linebreak\textit{e.g.}, ``the second package`                                              \\
impl\_anaphora     & Used when a slot type is not specifically mentionned
                   \linebreak \textit{e.g.}, ``What is the price for Rio?''...``And for Cleveland?''            \\
ref                & Id of the frame that the speaker is referring to                                  \\
read               & Reads slot values specified in another frame and writes them in the current frame \\
write              & Writes slot values in a given frame                                               \\
intent             & User intent (\textit{e.g.}, book)                                                 \\
action             & Wizard action (\textit{e.g.}, book)                                               \\
\bottomrule
\end{tabu}
\label{tab:slot_type}
\end{center}
\end{table}


\end{document}